\documentclass{article}

\PassOptionsToPackage{numbers,compress}{natbib}
\usepackage[dblblindworkshop,final]{neurips_2026}
\workshoptitle{AI for Drug Discovery (AI4DD)}

\usepackage[utf8]{inputenc}
\usepackage[T1]{fontenc}
\usepackage{hyperref}
\usepackage{url}
\usepackage{booktabs}
\usepackage{amsfonts}
\usepackage{amsmath}
\usepackage{amssymb}
\usepackage{multirow}
\usepackage{nicefrac}
\usepackage{microtype}
\usepackage{xcolor}
\usepackage{graphicx}
\usepackage{float}
\usepackage{threeparttable}

\newif\ifshowrevisions
\showrevisionsfalse
\newcommand{\rev}[1]{\ifshowrevisions{\color{red}#1}\else#1\fi}
\newcommand{\framework}{RangeFinder}

\title{RangeFinder: Predicting Ligand Property Ranges from Protein Pockets}

\author{%
  Alan Nadelsticher Ruvalcaba$^{1}$\thanks{Corresponding author.} \\
  Georgia Institute of Technology \\
  \texttt{alan.nadelsticher@gatech.edu}
  \And
  Chris Alvin$^{2}$ \\
  Furman University \\
  \texttt{chris.alvin@furman.edu}
  \AND
  Supratik Mukhopadhyay$^{3}$ \\
  Louisiana State University \\
  \texttt{mmukho1@lsu.edu}
  \And
  Adam Bess$^{3}$ \\
  Louisiana State University \\
  \texttt{bess@deepdrug.ai}
}

\begin{document}

\maketitle

\begin{abstract}
Identifying ligands for a protein target requires searching a vast chemical space for a comparatively small set of compatible molecules. Existing machine learning approaches primarily address this problem by generating candidate ligands or scoring individual protein--ligand pairs, but provide limited information about the broader chemical space compatible with a binding pocket. We introduce \framework, an SE(3)-equivariant transformer pocket-level model, trained and evaluated on BioLiP2 data, that predicts ranges over nine physicochemical ligand properties directly from pocket structure, chemistry, and evolutionary context. Using an intersection-over-union objective and evaluation metric (range-IoU), \framework~achieves a range-IoU of $0.464 \pm 0.002$, outperforming all deployable baselines while requiring only $12.2$ ms per pocket. We also quantify our method's ability to reduce the ligand search space while retaining compatible binders; in this regard, \framework~achieves a Reduction-Recall F-Score of $0.543 \pm 0.011$ --- once again outperforming all deployable baselines. Ablation experiments indicate that \framework's performance depends more strongly on the density and quality of ligand supervision than on any individual input modality or specialized training strategy. Together, these results position \framework~as an inexpensive and effective molecular filtering prior, with demonstrated potential in narrowing the ligand search space of target pockets before docking, generation, or synthesis. The code for this project can be found at: \url{https://github.com/alannadels/RangeFinder}.
\end{abstract}

\section{Introduction}

Identifying ligands for a protein target requires searching an enormous chemical space for a relatively small set of molecules that may bind a given pocket.
Two classes of machine learning methods have addressed this problem so far: generative models, which design candidate ligands directly, and scoring models, which rank existing candidates by predicted compatibility with the pocket.
Both have shown success within their respective problem spaces, yet both provide guidance at the level of individual molecules rather than describing the broader chemical space compatible with the pocket.

\framework~instead describes that space directly, predicting ranges over nine physicochemical ligand properties: molecular weight, lipophilicity, hydrogen-bond donors and acceptors, topological polar surface area, rotatable bonds, formal charge, aromatic rings, and fraction of sp$^3$-hybridized carbons, from pocket signal alone. 
Rather than proposing or evaluating individual molecules, the model describes the compatible chemical space through these ranges.
This description can focus subsequent generation or screening on candidate-like molecules.
Evaluated with the metrics shown in Sec.~\ref{sec:setup}, \framework~outperforms all deployable baselines on range-IoU and Reduction-Recall F-Score (RRF).
 
This work makes three contributions.
(1) It poses pocket-to-ligand modeling as a range-prediction task, as well as provides a corresponding metric suite and training and evaluation protocol.
(2) It shows empirically that \framework~outperforms all deployable baselines on range-IoU and RRF, effectively reducing the search space and time required by scientists to find viable binders for target pockets. (3) It examines, through controlled model variants and ablations, what drives this gain, where the current approach still falls short, and directions for future work.

\section{Related work}

Our task connects work on protein and pocket structure representation, structure-conditioned ligand generation, and pocket-ligand scoring.
Although the following literature addresses questions distinct from those posed here, it informed our methodological design and experimental process.

\noindent\textbf{Protein and pocket structure representation.}
APoc \citep{apoc} and Foldseek \citep{foldseek} are the two pipelines which most closely resemble \framework, and thus were directly used as baselines for our core experiments.
These methods compare binding sites or protein structures through geometric alignment and structural alphabets.
Other work in this space, such as AlphaFold 3 \citep{alphafold3}, extends structure prediction to complexes containing proteins and small molecules; however, it requires prior specification of the molecular components.
Furthermore, geometric deep-learning models, including SE(3)-equivariant networks \citep{se3transformer,e3nn}, and transformer-based protein language models, such as ESM-2 \citep{esm2}, can provide complementary learned representations of three-dimensional geometry and protein sequence, respectively.

\noindent\textbf{Structure-conditioned ligand generation.}
LiGAN \citep{ligan} is a structure-based generative model that proposes new ligand structures for a specified protein pocket. It represents the pocket and its potential ligand as three-dimensional density grids, allowing generated molecules to reflect the shape and chemical environment of the binding site. Similarly, 3D-SBDD \citep{3dsbdd} and Pocket2Mol \citep{pocket2mol} construct molecules atom by atom to satisfy the spatial and chemical constraints of a pocket. TargetDiff \citep{targetdiff} pursues the same goal through equivariant diffusion, while multi-objective methods optimize candidate ligand molecules across several desired properties \citep{multiobj}.

\noindent\textbf{Pocket--ligand scoring.}
RF-Score \citep{rfscore} predicts binding affinity from the types and numbers of atomic contacts between the protein and ligand, providing a learned alternative to traditional docking scores. AtomNet \citep{atomnet} and Pafnucy \citep{pafnucy} represent complexes on three-dimensional grids and use convolutional networks to learn spatial interaction patterns associated with activity or affinity. PotentialNet \citep{potentialnet} instead represents the complex as a graph and models both molecular structure and protein--ligand interactions. GNINA \citep{gnina} incorporates convolutional scoring into docking to rank candidate binding poses and ligands. 

Taken together, these methods represent or compare protein structures, predict molecular complexes, generate individual ligands, or score specified protein--ligand pairs.
None of them predict the physicochemical property space compatible with a pocket before a candidate molecule is proposed.

%which is the gap \framework~addresses.

\section{Method}
\label{sec:method}
\framework~is designed such that the predicted intervals depend only on the atomic chemistry, evolutionary context, and relative geometry of the pocket, and are invariant to global rotations and translations of its atomic coordinates.
%For each protein pocket, our framework predicts a range for each of nine physicochemical properties, summarizing the values of a pocket's unknown, potentially binding ligands. At the input level, the pocket is represented as a three-dimensional point cloud of protein atoms with chemical and evolutionary features, which are then encoded by an SE(3)-equivariant transformer. This design captures atomic chemistry and relative geometry, allowing the final prediction to be independent of the pocket's position or orientation.

\subsection{Task formulation}
\label{sec:task}
Given a protein pocket's 3D structure and its atom-level chemistry, our model's goal is to predict the range of nine physicochemical properties (Table~\ref{tab:properties}) spanned by the ligands that bind it.

\begin{figure}[H]
\centering
\includegraphics[width=\textwidth]{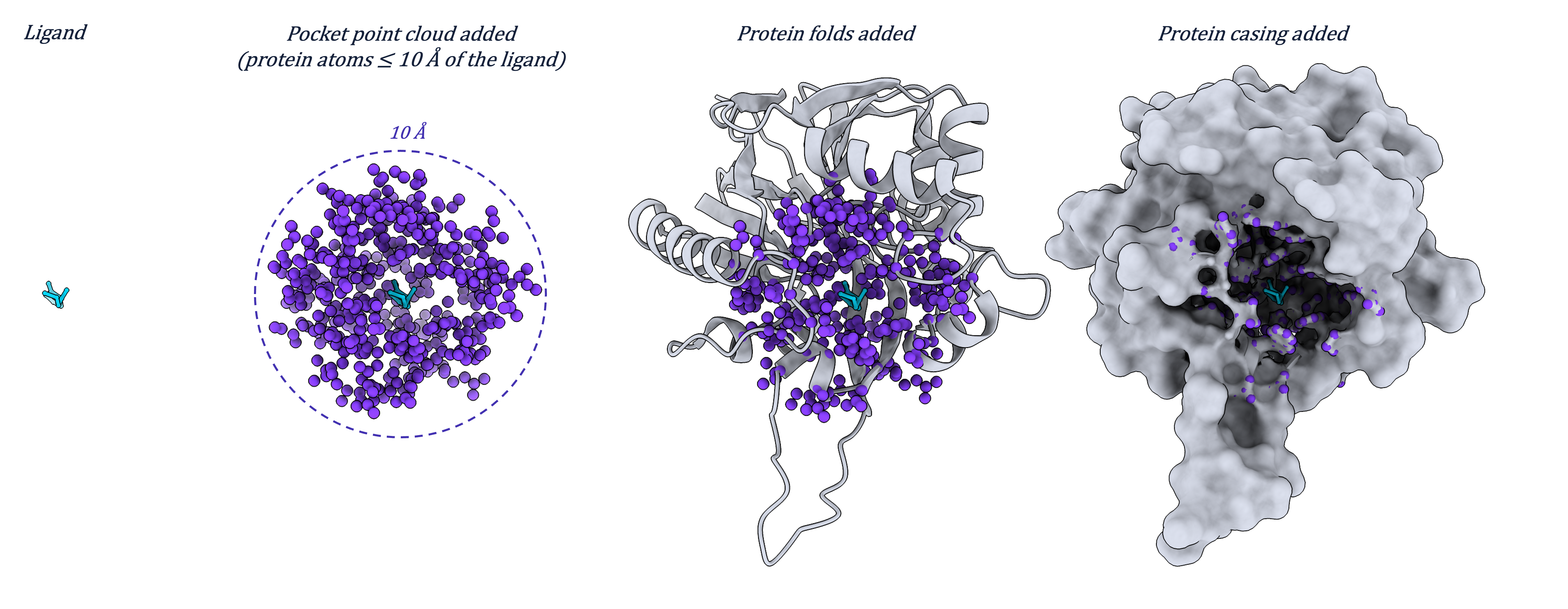}
\caption{Stepwise visualization of the structural pocket--ligand representation on which the model operates. The protein shown in the diagram, N-acyl-homoserine-lactone lactonase (PDB ID 6N9Q), represents a real example of a protein containing only one pocket as an active binding site.}
\label{fig:pocket-construction}
\end{figure}

\noindent\textbf{Input features.}
Each pocket is represented as the set of protein atoms within $10$ \AA{} of its ligand, with 3D coordinates being retained by each atom (Figure~\ref{fig:pocket-construction}). \rev{We adopt a $10$ \AA{} radius to capture the residue shells that line and shape the binding site while keeping the point cloud compact. During inference, for pockets with no known ligands, the $10$ \AA{} pocket for which the ranges are predicted is measured from whatever locates the binding site, for instance a reference molecule, a set of residues, or a point in space --- a bound drug-like ligand is not strictly required.} Every atom also carries a $1326$-d feature vector. It concatenates $46$ chemical and structural indicators, namely element, residue type, and bonding flags, with the $1280$-d ESM-2 \citep{esm2} embedding of its residue.
The coordinates supply geometry; the features supply chemistry and evolutionary context.

\noindent\textbf{Output predictions.}
Let $L_p$ denote the known ligands of pocket $p$.
For each property $j$ we compute the empirical mean $\mu_{pj}$ and standard deviation $\sigma_{pj}$ over $L_p$ and define the target interval $I_{pj} = [\mu_{pj}-2\sigma_{pj},\ \mu_{pj}+2\sigma_{pj}]$.
The model outputs $\hat{\mu}_{pj}$ and $\hat{\sigma}_{pj}$ for all nine properties, which define the predicted intervals $\hat{I}_{pj}$. \rev{A non-negative interval width is guaranteed by taking the absolute value of the predicted scale, $\hat{I}_{pj} = [\hat{\mu}_{pj} - 2|\hat{\sigma}_{pj}|,\ \hat{\mu}_{pj} + 2|\hat{\sigma}_{pj}|]$, applied identically in training and at inference.}
Importantly, we only use pockets with at least three known ligands for both training and evaluation, as pockets with fewer than three ligands lack a reliable range.

\subsection{Architectural pipeline}

The model maps a pocket to the interval parameters of Section~\ref{sec:task} in four stages: input embedding, equivariant attention, invariant pooling, and prediction.
The first three stages form the encoder, an SE(3)-Transformer \citep{se3transformer} implemented with \texttt{e3nn} \citep{e3nn}.
% Stages 1 and 2, which represent the very foundation of our model, are illustrated in figures \ref{fig:stage1} and \ref{fig:stage2}, respectively.

\begin{figure}[H]
\centering
\includegraphics[width=\textwidth]{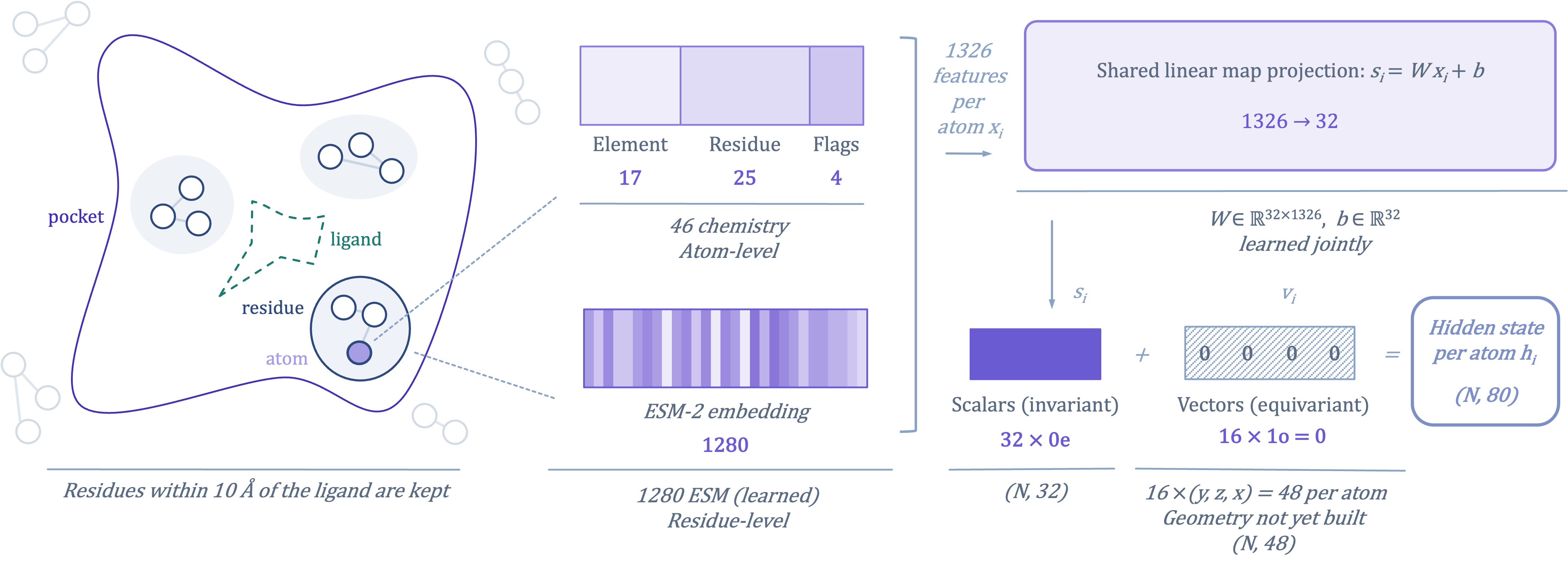}
\caption{Stage 1 -- Input embedding. Each pocket atom receives $46$ local chemistry and structural descriptors together with its residue's $1280$-dimensional ESM-2 embedding. The resulting $1326$-dimensional feature vector is projected to $32$ invariant scalar channels, while $16$ equivariant vector channels are initialized to zero.}
\label{fig:stage1}
\end{figure}

\noindent\textbf{(1) Input embedding.}
Each atom $i$ enters with $1326$ scalar, rotation-invariant features, $x_i \in \mathbb{R}^{1326}$.
A single linear layer maps each atom's feature vector to $32$ scalar channels, $s_i = W x_i$ with $W \in \mathbb{R}^{32 \times 1326}$.
Since the per-atom inputs carry no directional information, the $16$ vector channels are initialized to zero, $V_i = \mathbf{0} \in \mathbb{R}^{16 \times 3}$.
The atom's hidden state $h_i = (s_i, V_i)$ therefore lives in the irreducible representations $\underbrace{32 \times 0e}_{\text{invariant scalars}} \;+\; \underbrace{16 \times 1o}_{\text{equivariant vectors}}$.
That is, the hidden state consists of $32$ scalars that are unchanged under rotation and $16$ three-dimensional vectors of odd parity that rotate with the pocket.
Its width is therefore $80$ components per atom, $32 + 16 \times 3$.
Equivariant vector features are synthesized only in the following layers, from the relative positions of neighboring atoms. Figure~\ref{fig:stage1} illustrates this stage.

\begin{figure}[H]
\centering
\includegraphics[width=\textwidth]{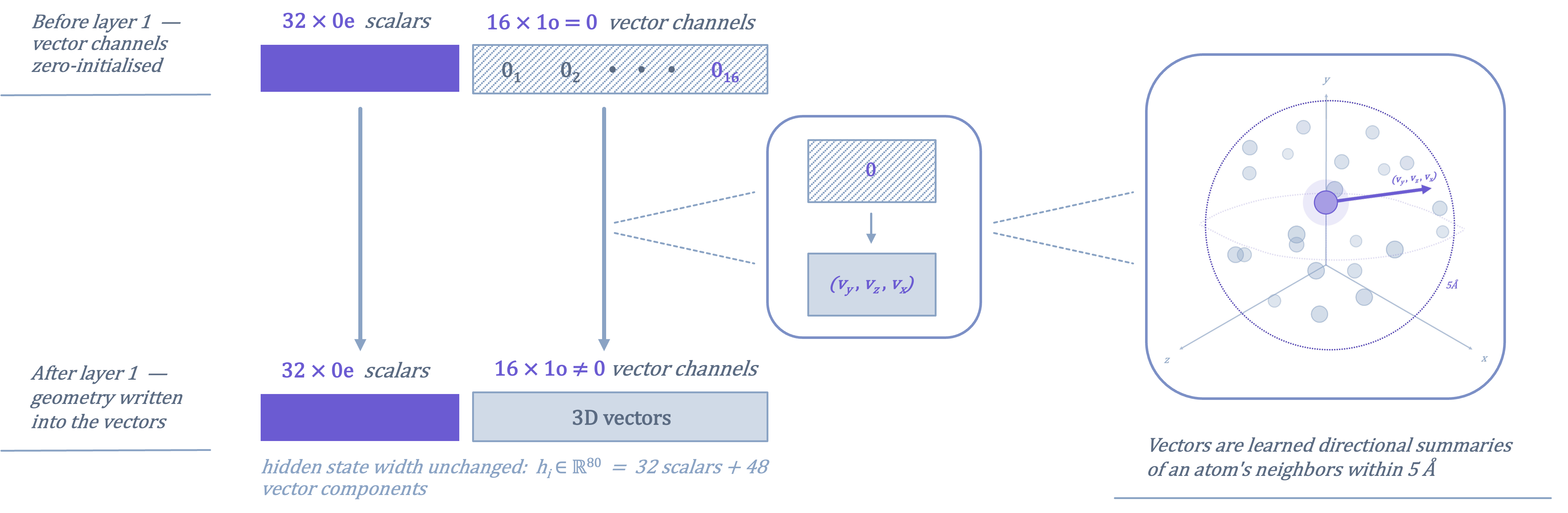}
\caption{Stage 2 -- Equivariant attention. The zero-initialized vector channels acquire directional information from neighboring atoms within $5$\,\AA{}, while the scalar channels remain rotation invariant. The hidden-state width remains at $80$ components per atom: $32$ scalars and $16$ three-dimensional vectors.}
\label{fig:stage2}
\end{figure}

\noindent\textbf{(2) Equivariant attention.}
As shown in Figure~\ref{fig:stage2}, four SE(3)-Transformer layers operate on a radius graph built from the pocket's atomic coordinates, connecting atoms within $5$\,\AA{} of each other.
Each edge is embedded by a Gaussian radial basis modulated by a smooth cosine cutoff, and its direction by spherical harmonics up to $\ell=1$.
Per layer, a tensor product combines each source atom's features with the edge harmonics to form messages. Lightweight $4$-head attention over the scalar channels weights those messages.
The messages are then aggregated at each atom and passed through a gated nonlinearity, a residual connection, and equivariant batch normalization.
Because messages are constructed from relative positions through equivariant tensor products, the encoder is equivariant to global rotations and translations of the pocket.

\noindent\textbf{(3) Invariant pooling.}
Atom representations are mean-pooled per pocket. The scalar channels and the norms of the vector channels form a $48$-dimensional rotation-invariant pocket embedding.

\noindent\textbf{(4) Prediction head.}
A multilayer perceptron with hidden dimensions $96$ and $48$ maps the pooled embedding to $18$ interval parameters, $\hat\mu$ and $\hat\sigma$ for each of the nine properties.
It also produces a single logit for an auxiliary binary task that predicts whether the pocket's ligands contain aromatic rings.

\subsection{Training objective}
The primary objective is a one-dimensional generalized-IoU (GIoU) loss \citep{giou} on the predicted and target intervals.
It is applied in raw property units: predictions are made in a normalized space and de-normalized before the intervals are formed.
For property $j$ of pocket $p$, let $U$ be the union length of $\hat I_{pj}$ and $I_{pj}$, and let $C$ be the length of the smallest interval enclosing both.
The loss is then $\ell_{pj} = 1 - \operatorname{GIoU}(\hat I_{pj}, I_{pj})$, where $\operatorname{GIoU} = \operatorname{IoU} - \frac{C - U}{C}$.
The enclosing-interval penalty $\tfrac{C-U}{C}$ supplies a nonzero gradient even when the two intervals are disjoint so the model is always pushed to move its interval toward the target.
Only properties with a non-degenerate target ($\sigma_{pj}>0$) contribute.
Furthermore, per-pocket losses are averaged over valid properties and then combined across pockets with cost-sensitive weights (inverse protein frequency), yielding $\mathcal{L}_{\mathrm{GIoU}}$, so that over-represented proteins do not dominate.
Because the GIoU term is computed on the same $\mu \pm 2\sigma$ intervals used at evaluation, it is a differentiable surrogate for the range-IoU metric.
The full objective adds a binary cross-entropy loss $\mathcal{L}_{\mathrm{BCE}}$ on the auxiliary head.
The total is $\mathcal{L} = \mathcal{L}_{\mathrm{GIoU}} + \lambda\,\mathcal{L}_{\mathrm{BCE}}$, where $\lambda = 0.089$ weights the auxiliary BCE loss relative to the primary objective, and was selected by CMA-ES search.

\section{Experiments}
\label{sec:exp}

\subsection{Experimental setup}
\label{sec:setup}
\noindent\textbf{Dataset.}
We train and evaluate our method on BioLiP2 \citep{biolip2}, a database that has biologically relevant protein--ligand interactions extracted from experimentally determined structures in the Protein Data Bank \citep{pdb}.
Table~\ref{tab:data} reports the training, validation, and test split.
We split the data by protein so that related structural views of the same protein cannot appear across training, validation, and test sets.
We evaluate all methods on the same $564$ test pockets, each with at least three known ligands.

\noindent\textbf{Baselines.}
We compare \framework~against six baselines. (1) \emph{Global average} averages the property ranges over all training pockets, giving no per-pocket signal.
(2) \emph{Pocket cluster} applies $K$-means \citep{kmeanspp} ($K = 6$) to pooled pocket features and predicts the mean target of the assigned training cluster, clustering at the pocket (input) level.
(3) \emph{Feature $k$-NN} \citep{coverhart} averages the true ranges of a pocket's $k$ nearest training pockets ($k = 50$).
(4) \emph{APoc} \citep{apoc} and (5) \emph{Foldseek} \citep{foldseek} perform the same nearest-neighbor retrieval as Feature $k$-NN, but rank training pockets by 3D structural similarity rather than feature similarity.
(6) \emph{Target-informed} applies $K$-means ($K = 6$) at the ligand (output) level, clustering pockets by their ground-truth labels; because it clusters on the prediction target itself, we include it as a non-deployable reference point.

\noindent\textbf{Evaluation metrics.}
We score a method's prediction via five metrics: \textit{Intersection-over-Union (range-IoU), Reduction, Recall, Reduction-Recall F-Score (RRF)}, and \textit{latency}. This metric suite provides a balanced evaluation, pairing predictive accuracy (range-IoU) with the practical screening trade-off between shrinking the candidate library (reduction) and retaining true binders (recall) --- summarized by the harmonic mean of recall and the library discard fraction (RRF) --- together with deployment cost (latency). A more in-depth explanation of what these metrics represent, and how they are computed, can be found in Section~\ref{sec:metrics} of the appendix.
% Predictions are evaluated with interval intersection-over-union defined as $\operatorname{IoU}(I,\hat I) = \frac{|I\cap\hat I|}{|I\cup\hat I|}$,
% and averaged first across the nine properties and then across pockets; we call this \emph{range-IoU}.
% Importantly, the metric rewards both accurate placement and appropriate width --- an overly broad interval cannot score highly simply by containing the target.

\rev{\noindent\textbf{Hyperparameters.} Several training hyperparameters were selected using a prior CMA-ES search \citep{cmaes} and held fixed across all subsequent experiments; Table \ref{tab:hyperparameters} in Appendix~\ref{sec:hyperparameters} provides additional search details. Full architecture, optimization, loss, and pocket-construction details are similarly provided in the aforementioned table.}
\rev{}

%\noindent\textbf{Hardware.}
% All main results are produced on a machine with an NVIDIA A10G GPU, used for both training and evaluation. The ablation models are trained on an NVIDIA RTX 5090 and evaluated on the A10G; every reported score uses the same evaluation hardware as the main model.
\noindent\textbf{Hardware.}
NVIDIA hardware was used for all experiments, and evaluation was conducted under identical hardware to ensure fairness. Additional details are shown in Section~\ref{sec:hardware}.

\subsection{Findings}
\label{sec:findings}

\noindent\textbf{Baseline comparison.}
As shown in Table~\ref{tab:main}, \framework~outperforms every deployable baseline tested on range-IoU and RRF, including the published methods Foldseek ($0.390$, 0.278) and APoc ($0.420$, 0.424). Compared to the best-performing baseline (APoc), on range-IoU \framework~exceeds their score by $0.044$, a $\sim 10.5\%$ relative improvement; on RRF, \framework~achieves a $28.1\%$ relative improvement over their result --- all while running approximately three orders of magnitude faster ($12.2$ ms vs.\ $26.3$ s per pocket). The global average, feature $k$-NN, and pocket clustering baselines do not approach this accuracy. Notably, \framework~nearly matches the non-deployable target-informed reference, which clusters pockets using the ligands' ground truth properties.

\noindent\textbf{Ablations.}
\label{sec:ablations}
We run a set of three ablation experiments.
The first isolates the contribution of the data source on which the model's pipeline runs, described further in Section~\ref{sec:ablation3} and Table~\ref{tab:supervision}. The second isolates the contribution of each input modality.
Starting from the full model, we remove the ESM-2 language-model embeddings (\emph{No ESM}), the hand-crafted chemical features (\emph{No chemistry}), and the 3D geometry (\emph{No 3D}), one at a time, and retrain (Table~\ref{tab:ablation}).
The third fixes the inputs and varies the training strategy and architecture.
We compare \framework~against a \emph{denoising-pretrained encoder} and a \emph{mixture-of-experts} approach routed over pocket clusters (Table~\ref{tab:ablation2}). For these experiments, we evaluate on the range-IoU and latency metrics previously introduced. A breakdown of results, across all ablation experiments, can be found in Section~\ref{sec:ablations_appendix}.
% This tells us whether self-supervised pretraining or cluster-specific signaling buys accuracy over \framework.

%indicating that the predictive signal is distributed and partially redundant across geometry, evolutionary, and chemical features rather than concentrated in any one of them.

\begin{table}[H]
    \centering
    \scriptsize
    \begin{threeparttable}
    \caption{Main experimental results on BioLiP2 data. \framework~was trained with seeds $42$--$46$ (five seeds); the other marked methods
  were run with seeds $0$--$4$ (five seeds). \textbf{Bold}/\underline{underline}: best/second-best per column, excluding the non-deployable
  reference.}
    \label{tab:main}
    \setlength{\tabcolsep}{2pt}
    \begin{tabular*}{\linewidth}{@{\extracolsep{\fill}}lccccc@{}}
    \toprule
    Method & Range-IoU & Reduction & Recall & RRF & Latency \\
    \midrule
    Global average        & 0.402 & $3.0\times$ & 0.205 & 0.313 & \textbf{\boldmath$<0.001$\,ms} \\
    Foldseek              & 0.390 & \underline{$4.2\times$} & 0.170 & 0.278 & $127$\,ms \\
    Pocket cluster\tnote{*} & $0.403 \pm 0.000$ & $3.3\times$ & $0.222 \pm 0.004$ & $0.338 \pm 0.004$ & \underline{$2.83$\,ms} \\
    Feature $k$-NN\tnote{*} & $0.409 \pm 0.000$ & $4.1\times$ & $0.214 \pm 0.002$ & $0.333 \pm 0.002$ & $3.01$\,ms \\
    APoc                  & \underline{0.420} & \textbf{\boldmath$5.3\times$} & \underline{0.287} & \underline{0.424} & $26.3$\,s \\
    \framework~(Ours)\tnote{*} & \textbf{\boldmath$0.464 \pm 0.002$} & $2.1\times$ & \textbf{\boldmath$0.557 \pm 0.026$} & \textbf{\boldmath$0.543
  \pm 0.011$} & $12.2$\,ms \\
    \emph{Target-informed}\tnote{*} & \emph{$0.470 \pm 0.005$} & \emph{$3.8\times$} & \emph{$0.510 \pm 0.011$} & \emph{$0.602 \pm 0.012$} &
  \emph{$0.195$\,ms} \\
    \bottomrule
    \end{tabular*}
    \begin{tablenotes}
    \footnotesize
    \item[*] Marked methods contain tunable stochastic components.
    \end{tablenotes}
    \end{threeparttable}
  \end{table}

\section{Discussion}
\label{sec:discussion}
As shown in Section~\ref{sec:findings}, \framework~provides the strongest prediction (range-IoU: $0.464 \pm 0.002$) and Reduction-Recall F-Score (RRF: $0.543 \pm 0.011$), outperforming all deployable baselines.
It also nearly matches the non-deployable, target-informed baseline on range-IoU prediction.
Furthermore, at roughly $12$\,ms per pocket, its inference latency is substantially lower than APoc's and Foldseek's.
All methods, including the non-deployable baseline, nonetheless fall within $\sim 0.08$ range-IoU of one another, and the pocket-blind global average alone stands at $0.402$.
This narrow spread suggests that the available signal is modest, and that sparse, noisy data rather than model capacity may be the binding constraint.
Ultimately, \framework~provides an inexpensive property-space prior that could filter virtual libraries, constrain molecular generation, or triage pockets before further molecular analysis and testing.

Furthermore, removing any single input modality reduces range-IoU by at most $0.012$, and neither mixture-of-experts training nor denoising pretraining improves on \framework.
Together with the supervision-source ablation experiment, this indicates that performance is not attributable to any single input modality or training strategy. As previously noted, the larger determinant is the density and reliability of the ligand sets that define each pocket's target ranges during training.

%However, the narrow overall performance band suggests that the available signal remains modest and may be constrained by sparse, noisy empirical ranges rather than model capacity alone.

\noindent\textbf{Limitations.}
\rev{Key limitations include evaluation on $3+$ ligand pockets and noisy range estimates near the three-ligand threshold, uncertainty in small evaluation margins from single-seeded splits and ablations, and possible homology or pocket-size confounds. A detailed discussion is provided in Appendix \ref{sec:limitations}.}
% Despite our strong results, some limitations remain. First, the predicted intervals represent a coarse decision aid rather than a direct substitute for generation, affinity prediction, or experimental validation. Second, our model and the baselines use one fixed seed (seed=$42$) for evaluation, training (where applicable), and the protein-grouped data split; thus, results within small margins, especially the $0.001$ gap over the target-informed reference, should not be over-interpreted.

\noindent\textbf{Future work.}
\rev{Future work aims to close the gaps raised by the limitations section, and explore potentially fruitful research directions; such work is discussed in Appendix \ref{sec:future_work}.}

% Future research should address the limitations previously raised, and explore other promising directions. Concretely, subsequent work should: (1) explore complementary methods or evaluation protocols that, when combined with our framework, can provide more robust pocket-ligand predictions; (2) rerun our experimental pipeline across multiple seeds and data splits; (3) extend the task to pockets with cofactors, where these cofactors could themselves be used as input features; and (4) formulate a testing ground to determine whether our model's predicted ranges exclude true binders, and how to remedy this behavior if observed.

\section{Conclusion}
We introduced \framework, a pocket-level model that, from pocket geometry, chemistry, and evolutionary context predicts the range of nine physicochemical properties spanned by a pocket's potential binders.
On BioLiP2 it achieves $0.464 \pm 0.002$ range-IoU and $0.543 \pm 0.011$ RRF, outperforming all deployable baselines while requiring only $12.2$\,ms per pocket.
Furthermore, our input and training-strategy ablations indicate that this result is not explained by feature or structural retrieval, by any single input modality, or by specialized optimization.
A major contributing factor to the model's performance, rather, seems to be data density and quality; over its own geometry-blind baseline, a PDBbind~\citep{pdbbind} \framework~variant improves by only $+0.006$, compared to $+0.064$ for the BioLiP2 variant (both at seed $=$ $42$). Together, these results illustrate the major drivers behind \framework's~performance, and position it as an inexpensive and promising method, with the potential for narrowing the ligand search space of target pockets before docking, generation, or synthesis.
%Because prediction requires only pocket structure, it applies in principle to orphan pockets with no known binders, a regime this work does not yet evaluate. 

%\clearpage
%\medskip

\appendix

\section*{Appendix}
The appendix section contains additional details, results, and figures that provide further context regarding our work.

\section{Task and model details}

This section provides additional details on the ligand properties predicted by \framework, the construction of its pocket representations, and the model's input embedding and equivariant attention stages.

\subsection{Predicted ligand properties}
\framework~predicts ranges over nine physicochemical properties --- the targeted ligand property space associated with each pocket. Table~\ref{tab:properties} lists these properties and their definitions.

\begin{table}[H]
\centering
\small
\begin{threeparttable}
\caption{The nine ligand physicochemical properties predicted by \framework.}
\label{tab:properties}
\begin{tabular}{ll}
\toprule
Property & Definition \\
\midrule
MW & Molecular weight \\
logP & Octanol--water partition coefficient (lipophilicity) \\
HBD & Number of hydrogen-bond donors \\
HBA & Number of hydrogen-bond acceptors \\
TPSA & Topological polar surface area \\
Rotatable bonds & Number of rotatable bonds \\
Formal charge\tnote{*} & Net formal charge \\
Aromatic rings & Number of aromatic rings \\
Fsp$^3$ & Fraction of sp$^3$-hybridized carbons \\
\bottomrule
\end{tabular}
\begin{tablenotes}[flushleft]
\footnotesize
\item[*] In $81\%$ of test pockets, $\sigma_{\mathrm{charge}}<0.1$, making the target interval nearly degenerate (discussed below).
\end{tablenotes}
\end{threeparttable}
\end{table}

\paragraph{Formal charge as a degenerate interval target.}
Formal charge behaves differently from the other properties. In $81\%$ of the test pockets, $\sigma_{\mathrm{charge}}<0.1$, so the target interval is effectively a point near zero. Any small continuous prediction error can then produce zero overlap. Averaged over the five seeds, \framework's~predictions for formal charge achieve a range-IoU of $0.124$, whereas those for the other eight properties lie between $0.485$ and $0.527$ (mean $0.506$). Dropping formal charge, \framework's~8-property range-IoU is $0.506$ --- computed as $(0.464\times 9 - 0.124)/8$ --- a range-IoU improvement of $0.042$ ($\sim 9.1\%$) relative to the 9-property score of $0.464$.

\subsection{Pocket construction}

\subsection{Architecture visualizations}

\section{Dataset details}

\begin{table}[H]
\centering
\small
\caption{BioLiP2-based dataset split by protein. A structure-view is one observed protein--ligand entry associated with a pocket.}
\label{tab:data}
\begin{tabular}{lccc}
\toprule
 & Train & Validation & Test \\
\midrule
Pockets ($n_{\mathrm{lig}}\geq3$) & 2{,}627 & 562 & 564 \\
Structure-views                    & 10{,}905 & 2{,}313 & 3{,}823 \\
\bottomrule
\end{tabular}
\end{table}

\section{Evaluation metrics}
\label{sec:metrics}
Below is a detailed explanation of what each metric in our suite represents, and how it is computed.

\noindent\textbf{Intersection-over-union.} We define intersection-over-union as the
overlap between a predicted property interval and its target relative to their union,
where $\operatorname{IoU}(\hat I_{pj}, I_{pj}) = \frac{|\hat I_{pj} \cap I_{pj}|}{|\hat I_{pj} \cup I_{pj}|}$,
averaged across the nine properties and then across pockets; we call this
\emph{range-IoU}. This metric rewards both accurate placement and appropriate width,
since an overly broad interval cannot score highly simply by containing the target.

\noindent\textbf{Reduction.} We define reduction as $1/r_p$ averaged across pockets,
where $r_p$ is the fraction of a fixed background library $\mathcal{B}$ (all drug-like
binders in the dataset), whose nine properties fall within pocket $p$'s predicted box.
Reduction is the factor by which the predicted ranges shrink the candidate set relative
to screening the full library; larger values indicate a more selective filter. On its
own it can be misleading, since a very narrow box maximizes it but might also be discarding true binders; thus, it should be paired with, and read alongside recall.

\noindent\textbf{Recall.} We define recall as the fraction of a pocket's known binders that fall within its
predicted intervals, averaged across pockets. A binder counts as captured when each of its
nine property values lies within the corresponding predicted interval $\hat I_{pj}$. Recall measures how well the predicted ranges retain true binders; intervals
that are too narrow or misplaced which discard real ligands are penalized.

\noindent\textbf{Reduction-Recall F-Score.} We define RRF as the harmonic mean of
recall and the discard fraction $1-r_p$, $\;\mathrm{RRF} = \frac{2\,\text{recall}\,(1-r_p)}{\text{recall} + (1-r_p)}$,
averaged across pockets. It collapses the recall--reduction trade-off into a single
number, rewarding methods that both retain binders and shrink the library. We use the
harmonic mean rather than a sum so that a large reduction cannot compensate for lost
binders: a method scores highly only when \emph{both} quantities are high.

\noindent\textbf{Latency.} We define latency as the mean wall-clock time to produce a
prediction for a single pocket during inference, measured on identical hardware across all methods. It
captures deployability: methods of comparable accuracy can differ by orders of magnitude
in computational cost.

\rev{
\section{Training and search configuration}
\label{sec:hyperparameters}
Table~\ref{tab:hyperparameters} reports the full configuration of the deployed model.

\begin{table}[t]
\rev{
\centering
\small
\caption{RangeFinder training, loss, pocket-construction, and CMA-ES search configuration.}
\label{tab:hyperparameters}
\begin{tabular}{@{}l p{0.60\linewidth}@{}}
\toprule
Setting & Value \\
\midrule
\multicolumn{2}{@{}l}{\emph{Architecture}} \\
Encoder                       & SE(3)-Transformer (\texttt{e3nn}); $4$ layers, $4$ heads, $\ell_{\max}=1$ \\
Hidden irreducible reps. (width)         & $32\times0e + 16\times1o$ ($80$ components per atom) \\
Radius graph / radial basis   & $5$~\AA{} cutoff; $16$ Gaussian radial basis functions \\
Prediction head               & MLP $96\!\to\!48\!\to\!19$ ($18$ interval parameters $+1$ aromatic logit) \\
\midrule
\multicolumn{2}{@{}l}{\emph{Input features and normalization}} \\
Per-atom features             & $46$ chemical/structural $+$ $1280$ ESM-2 $= 1326$ \\
ESM-2 checkpoint              & \texttt{esm2\_t33\_650M\_UR50D} ($1280$-d residue embedding) \\
Coordinates                   & centroid-centered 3D positions (\AA{}) \\
Target normalization          & per-target $z$-score on training pockets; $\mu$ over all pockets, $\sigma$ over $\geq\!3$-ligand pockets \\
\midrule
\multicolumn{2}{@{}l}{\emph{Optimization}} \\
Optimizer                     & AdamW \\
Learning rate                 & $2.23\times10^{-4}$ \\
Weight decay                  & $1\times10^{-4}$ \\
LR schedule                   & ReduceLROnPlateau; factor $0.5$, patience $10$ \\
Batching                      & dynamic atom-budget, $\leq\!80{,}000$ graph edges per batch \\
Dropout (head / trunk / atom) & $0.245$ / $0.235$ / $0.0$ \\
Gradient clipping (max norm)  & $3.97$ \\
Training duration             & early stopping on validation (patience $12$); LR reduced on plateau (patience $10$) \\
Seeds                          & Five-seed bands: $42$--$46$ \\
\midrule
\multicolumn{2}{@{}l}{\emph{Loss and interval parameterization}} \\
Primary loss                  & $1$-D generalized-IoU on $\mu\pm2\sigma$ intervals, inverse-protein-frequency weighted \\
Auxiliary loss                & aromatic-present BCE, weight $\lambda = 0.089$ \\
Positive-width parameterization & $\hat{I}_{pj} = [\hat\mu_{pj}-2|\hat\sigma_{pj}|,\ \hat\mu_{pj}+2|\hat\sigma_{pj}|]$ \\
\midrule
\multicolumn{2}{@{}l}{\emph{Pocket and target construction}} \\
Pocket grouping               & by UniProt accession; target $\mu,\sigma$ computed over all drug-like ligands of the protein ($\geq\!3$ required) \\
Input point cloud             & protein atoms within $10$~\AA{} of a bound ligand (one representative structure per pocket) \\
\midrule
\multicolumn{2}{@{}l}{\emph{CMA-ES search}} \\
Details & early-stage search on PDBbind data; objective: validation Spearman; held fixed thereafter\\ Hyperparameter search & selected the learning rate, weight decay, dropouts (head, trunk, and atom), and gradient clip; population size $8$, $6$ generations ($48$ evaluations),
initial step size $\sigma = 0.5$ \\
Loss-weight search    & selected the auxiliary BCE weight $\lambda = 0.089$; population size $8$, $6$ generations ($48$ evaluations), initial step size $\sigma = 0.5$ \\
\bottomrule
\end{tabular}
}
\end{table}}

\section{Hardware}
\label{sec:hardware}
Unless otherwise noted, training and evaluation use NVIDIA A10G GPUs. The main results in Table~\ref{tab:main} are both trained and evaluated on these cards.

The supervision-source variants in Table~\ref{tab:supervision} are likewise trained and evaluated entirely on the A10G GPUs (the BioLiP2 variant being the same seed-$42$ model reported in Table~\ref{tab:main}). In contrast, the input-modality and
training-strategy ablations in Tables~\ref{tab:ablation} and~\ref{tab:ablation2} are trained on an NVIDIA RTX 5090 GPU and evaluated on the A10G cards.

To ensure fairness, all experiments were evaluated under the A10G GPUs --- regardless of training modality. This is particularly important for the latency metric which is heavily influenced by specific hardware.

\section{Additional ablations}
\label{sec:ablations_appendix}
For the first experiment (outlined in Section~\ref{sec:findings}), we observe that data quality and density have a strong impact on model performance; relative to their geometry-blind baselines, a PDBbind \framework~variant improves by only $+0.006$, compared to $+0.064$ for the BioLiP2 variant (both at seed $=$ $42$). Next, among the input modalities, removing the 3D geometry causes the most meaningful drop in performance.
Yet, this drop is modest, from $0.467$ to $0.455$.
Removing the ESM-2 embeddings or the chemical features reduces range-IoU by only $0.003$.
No single input is therefore critical, and the model degrades gracefully under the loss of any one modality.
In the third experiment, we find that neither the mixture-of-experts ($0.442$) nor the denoising-pretrained encoder ($0.461$) improves on \framework~($0.467$) on range-IoU.
Per-cluster specialization and self-supervised pretraining, therefore, offer no benefit over end-to-end training at this scale.

\subsection{Supervision source: BioLiP2 vs. PDBbind}
\label{sec:ablation3}
We ran the identical \framework~model --- same architecture, features, loss, and
training protocol --- under two supervision sources, and evaluated each model on its own
protein-grouped test split (Table~\ref{tab:supervision}). Because the two splits contain different pockets, the
absolute scores are not directly comparable; the meaningful quantity is each
model's margin over its same-dataset baselines. Under PDBbind, where most pockets
have only a few known binders, a geometry-blind pocket-cluster constant 
already nearly matches \framework~($0.431$ vs. $0.437$, respectively), leaving a margin of just $+0.006$ in favor of \framework.
Under BioLiP2, which enriches each pocket's ligand set by roughly an order of
magnitude, \framework~rises to $0.467$ while the geometry-blind pocket-cluster constant drops to $0.403$ --- a $+0.064$ margin in favor of \framework. The global-average floor is essentially
unchanged between the two ($0.403$ for PDBbind vs. $0.402$ for BioLiP2), so the growth in margin is
attributable to denser, sharper supervision rather than to an easier test set.

\begin{table}[H]
  \centering
  \small
  \caption{Range-IoU under two supervision sources, with the model held fixed. Each column uses its own test split; compare each model’s margin over its same-dataset baselines. \framework~variants were trained under the same fixed seed (seed $=42$).}
  \label{tab:supervision}
  \begin{tabular}{lcc}
  \toprule
   & BioLiP2 & PDBbind \\
  \midrule  
  Global-average floor                       & 0.402 & 0.403 \\
  Pocket cluster $K{=}6$ (geometry-blind)     & 0.403 & 0.431 \\
  \framework                        & 0.467 & 0.437 \\
  \midrule  
  \framework~gain over geometry-blind baseline           & +0.064 & +0.006 \\
  \bottomrule
  \end{tabular}
  \end{table}

\subsection{Input modalities}

\begin{table}[H]
  \centering
  \small
  \caption{Leave-one-out input ablation. All other training and evaluation choices are held fixed. All ablation variants shown were trained under the same fixed seed (seed $=42$). \textbf{Bold}/\underline{underline}: best/second-best per column.}
  \label{tab:ablation}
  \begin{tabular}{lcccccc}
  \toprule
  Variant & ESM & Chemistry & 3D & Feature dim. & Range-IoU & Latency \\
  \midrule
  Full model & \checkmark & \checkmark & \checkmark & 1326 & \textbf{0.467} & \underline{12.2\,ms} \\
  No ESM              & ---        & \checkmark & \checkmark & 46   & \underline{0.464} & \underline{12.2\,ms} \\
  No chemistry        & \checkmark & ---        & \checkmark & 1280 & \underline{0.464} & \underline{12.2\,ms} \\
  No 3D               & \checkmark & \checkmark & ---        & 1326 & 0.455 & \textbf{6.8\,ms} \\
  \bottomrule
  \end{tabular}
  \end{table}

\subsection{Training strategies and architectures}

\begin{table}[H]
  \centering
  \small
  \caption{Training-strategy ablation. Alternative training strategies and architectures for \framework~on the same $564$ BioLiP2 test pockets. All ablation variants shown were trained under the same fixed seed (seed $=42$). \textbf{Bold}/\underline{underline}: best/second-best per column.}
  \label{tab:ablation2}
  \begin{tabular}{llcc}
  \toprule  
  Variant & Strategy & Range-IoU & Latency \\
  \midrule
  \framework           & Single model                & \textbf{0.467} & \textbf{12.2\,ms} \\
  Denoising-pretrained encoder   & Self-supervised pretraining & \underline{0.461} & \textbf{12.2\,ms} \\
  Mixture of experts ($K{=}6$)   & Per-cluster specialization  & 0.442 & \underline{15\,ms} \\
  \bottomrule
  \end{tabular}
  \end{table}

\section{Limitations}
\label{sec:limitations}
Training and evaluation are constrained to pockets with at least three known ligands, as pockets with fewer than three binders do not provide reliable range estimates. Thus, the current dataset may represent a biased sample of protein pockets, lacking undocumented or minimally documented pockets. Evaluation could, therefore, be restricted to comparatively well-characterized pockets, with performance on ligand-poor pockets remaining untested. \rev{Moreover, near the three-ligand floor, $\mu\pm2\sigma$ estimates may remain somewhat noisy.} Additionally, although five-seed bands are provided in the main results for all methods that introduce tunable stochastic components as part of their architecture, the single-seeded data pipeline split (train/val/test) and the single-seeded training for the ablation studies remain a limitation to be explored in future work. \rev{Split groups are determined by UniProt rather than by sequence-identity clusters, so close homologs may appear across splits; however, the small effect of removing ESM-2 (as shown in Table~\ref{tab:ablation}) indicates that sequence-embedding leakage is not a major driver of model performance. The current suite of training-strategy ablations (Table~\ref{tab:ablation2}) does not determine whether \framework's equivariant architecture is responsible for its performance; an additional non-equivariant version of \framework~would answer this definitively. Lastly, because pockets are typically delimited using the bound ligand, pocket size slightly varies depending on ligand size; thus, size-related properties such as molecular weight could be partially inferable from pocket dimensions rather than pocket chemistry, we do not fully disentangle this and leave it for future work.}

\section{Future work}
\label{sec:future_work}
Because supervision rather than architecture appears to be the binding constraint, the most promising direction lies in experimenting with denser, debiased, uncertainty-aware, and higher-quality pocket-ligand data. Beyond this, subsequent work should extend the task to pockets with few or no known ligands, \rev{rerun the pipeline across multiple UniProt data-split seeds and homology-clustered splits, and for ablations specifically across multiple training seeds}, extend the task to pockets with cofactors, using the cofactors themselves as input features, \rev{further ablate \framework's architecture by testing non-equivariant versions of it, and determine the extent, if any, to which the model's learning is affected by the delimited pocket size across ligands of varying proportions}. \rev{Additionally, the relatively small gap across the distinct leave-one-out models motivates a comparison with simpler variants of \framework; potentially, a simpler and cheaper variant could achieve comparable results. Moreover, assessing RangeFinder's added value for molecular generation could be worthwhile. This can be done by conditioning a simple ligand generative model on \framework's predicted property ranges (or its learned pocket embedding) and comparing against that same model without the conditioning, holding the base model, pocket inputs, and generation budget fixed. Lastly, future work should explore complementary methods or protocols that, when combined with our framework (or aspects of it), can provide more robust pocket-ligand predictions.}


{
\small
\begin{thebibliography}{99}

\bibitem{apoc}
Gao, M.\ \& Skolnick, J.\ (2013) APoc: large-scale identification of similar protein pockets. \emph{Bioinformatics} 29(5):597--604.

\bibitem{foldseek}
van Kempen, M.,\ et al.\ (2024) Fast and accurate protein structure search with Foldseek. \emph{Nature Biotechnology} 42:243--246.

\bibitem{alphafold3}
Abramson, J.,\ et al.\ (2024) Accurate structure prediction of biomolecular interactions with AlphaFold 3. \emph{Nature} 630:493--500.

\bibitem{se3transformer}
Fuchs, F. B., Worrall, D. E.,\ et al.\ (2020) SE(3)-Transformers: 3D roto-translation equivariant attention networks. In \emph{Proceedings of the Conference on Neural Information Processing Systems (NeurIPS)}.

\bibitem{e3nn}
Geiger, M.\ \& Smidt, T.\ (2022) e3nn: Euclidean neural networks. \emph{arXiv:2207.09453}.

\bibitem{esm2}
Lin, Z.,\ et al.\ (2023) Evolutionary-scale prediction of atomic-level protein structure with a language model. \emph{Science} 379(6637):1123--1130.

\bibitem{ligan}
Ragoza, M., Masuda, T.\ \& Koes, D. R.\ (2022) Generating 3D molecules conditional on receptor binding sites with deep generative models. \emph{Chemical Science} 13(9):2701--2713.

\bibitem{3dsbdd}
Luo, S.,\ et al. \ (2021) A 3D generative model for structure-based drug design. In \emph{Proceedings of the Conference on Neural Information Processing Systems (NeurIPS)}.

\bibitem{pocket2mol}
Peng, X.,\ et al.\ (2022) Pocket2Mol: Efficient molecular sampling based on 3D protein pockets. In \emph{Proceedings of the International Conference on Machine Learning (ICML)}.

\bibitem{targetdiff}
Guan, J., Qian, W. W.,\ et al.\ (2023) 3D equivariant diffusion for target-aware molecule generation and affinity prediction. In \emph{Proceedings of the International Conference on Learning Representations (ICLR)}.

\bibitem{multiobj}
Jain, M.,\ et al.\ (2023) Multi-objective GFlowNets. In \emph{Proceedings of the International Conference on Machine Learning (ICML)}.

\bibitem{rfscore}
Ballester, P. J.\ \& Mitchell, J. B. O.\ (2010) A machine learning approach to predicting protein--ligand binding affinity with applications to molecular docking. \emph{Bioinformatics} 26(9):1169--1175.

\bibitem{atomnet}
Wallach, I., Dzamba, M.,\ \& Heifets, A.\ (2015) AtomNet: A deep convolutional neural network for bioactivity prediction in structure-based drug discovery. \emph{arXiv:1510.02855}.

\bibitem{pafnucy}
Stepniewska-Dziubinska, M. M., Zielenkiewicz, P.\ \& Siedlecki, P.\ (2018) Development and evaluation of a deep learning model for protein--ligand binding affinity prediction. \emph{Bioinformatics} 34(21):3666--3674.

\bibitem{potentialnet}
Feinberg, E. N.,\ et al.\ (2018) PotentialNet for molecular property prediction. \emph{ACS Central Science} 4(11):1520--1530.

\bibitem{gnina}
McNutt, A. T.,\ et al.\ (2021) GNINA 1.0: Molecular docking with deep learning. \emph{Journal of Cheminformatics} 13(1):43.

\bibitem{giou}
Rezatofighi, H.,\ et al.\ (2019) Generalized intersection over union: A metric and a loss for bounding box regression. In \emph{Proceedings of the IEEE/CVF Conference on Computer Vision and Pattern Recognition}.

\bibitem{biolip2}
Zhang, C.,\ et al.\ (2024) BioLiP2: An updated structure database for biologically relevant ligand--protein interactions. \emph{Nucleic Acids Research} 52(D1):D404--D412.

\bibitem{pdb}
wwPDB Consortium. (2019) Protein Data Bank: The single global archive for 3D macromolecular structure data. \emph{Nucleic Acids Research} 47(D1):D520--D528.

\bibitem{kmeanspp}
Arthur, D.\ \& Vassilvitskii, S.\ (2007) $k$-means++: The advantages of careful seeding. In \emph{Proceedings of the Eighteenth Annual ACM-SIAM Symposium on Discrete Algorithms (SODA)}, pp.\ 1027--1035.

\bibitem{coverhart}
Cover, T.\ \& Hart, P.\ (1967) Nearest neighbor pattern classification. \emph{IEEE Transactions on Information Theory} 13(1):21--27.

\bibitem{cmaes}
Hansen, N. (2023) The CMA Evolution Strategy: A Tutorial. \emph{arXiv preprint arXiv:1604.00772}.

\bibitem{pdbbind}
Liu, Z.,\ et al.\ (2017) Forging the basis for developing protein--ligand interaction scoring functions. \emph{Accounts of Chemical Research} 50(2):302--309.

\end{thebibliography}
}
\end{document}